\documentclass[runningheads]{llncs}
\usepackage[T1]{fontenc}
\usepackage{graphicx}
\usepackage{amsmath,amssymb,amsfonts}
\usepackage{hyperref}
\usepackage{float}

\begin{document}
\title{FlowExtract: Procedural Knowledge Extraction from Maintenance Flowcharts}

\titlerunning{FlowExtract: Procedural Knowledge Extraction}

\author{Guillermo Gil de Avalle\inst{1}\orcidID{0009-0004-1538-2256} \and
Laura Maruster\inst{1}\orcidID{0000-0002-6588-7648} \and
Eric Sloot\inst{2}\orcidID{0009-0009-2419-6371} \and
Christos Emmanouilidis\inst{1}\orcidID{0000-0003-4335-6915}}
\authorrunning{G. Gil de Avalle et al.}
\institute{University of Groningen, Nettelbosje 2, Groningen, The Netherlands\\
\email{\{g.gil.de.avalle,l.maruster,c.emmanouilidis\}@rug.nl}
\and
Philips Consumer Lifestyle B.V., Oliemolenstraat 5, Drachten, The Netherlands
\\
\email{eric.sloot@philips.com}}
\maketitle
\begin{abstract}
Maintenance procedures in manufacturing facilities are often documented as flowcharts in static PDFs or scanned images. They encode procedural knowledge essential for asset lifecycle management, yet inaccessible to modern operator support systems. Vision-language models, the dominant paradigm for image understanding, struggle to reconstruct connection topology from such diagrams. We present \textit{FlowExtract}, a pipeline for extracting directed graphs from ISO 5807-standardized flowcharts. The system separates element detection from connectivity reconstruction, using YOLOv8 and EasyOCR for standard domain-aligned node detection and text extraction, combined with a novel edge detection method that analyzes arrowhead orientations and traces connecting lines backward to source nodes. Evaluated on industrial troubleshooting guides, FlowExtract achieves very high node detection and substantially outperforms vision-language model baselines on edge extraction, offering organizations a practical path toward queryable procedural knowledge representations. The implementation is available at \url{https://github.com/guille-gil/FlowExtract}.

\end{abstract}
\section{Introduction}

Flowcharts encode procedural knowledge as step-by-step instructions representing decision logic and action sequences, in a visual format that humans parse effortlessly. We trace arrows between boxes, interpreting shapes as actions or decisions and connections as control flow. This understanding relies on both semantic interpretation of node content and spatial reasoning about how elements relate~\cite{spatialvlm}. Automated systems face a harder task: parsing flowcharts requires text recognition, identification of visual elements, and reconstruction of the connections between them~\cite{flowlearn}. Although standards such as ISO 5807~\cite{iso5807} define unambiguous symbol conventions, procedural documentation often remains as scanned images or static PDF files, posing challenges for organizations seeking to extract the knowledge encoded within~\cite{mentzas2024}.

This challenge is particularly acute for asset lifecycle management. Maintenance procedures guide operators and technicians through equipment diagnostics, corrective actions, or preventive routines, encoding operational knowledge developed over decades of experience~\cite{proceduralkg}. However, this knowledge typically remains trapped in vast libraries of unorganized documentation that staff must manually navigate~\cite{mentzas2024}. Converting these procedures into queryable knowledge representations would enable intelligent decision support~\cite{emmanouilidis2019}, but manual digitization is prohibitively expensive at scale~\cite{proceduralkg}, creating demand for automated extraction methods that can reliably parse the flowchart structure.

Vision-language models (VLMs) have emerged as the dominant paradigm for joint document and image understanding~\cite{layoutlmv3}, encoding visual features and textual information through transformer architectures. VLMs demonstrate strong visual analysis capabilities, and can even interpret basic diagrams~\cite{flowlearn}. However, when applied to flowcharts with dense layouts, overlapping arrows, and/or complex branching structures, performance suffers substantially. Systematic evaluation reveals that while VLMs handle node and text detection reasonably well, they consistently fail to reconstruct the connection topology~\cite{flowlearn}. Pipeline approaches combining object detection with language models face similar limitations, as converting spatial layouts into textual descriptions discards positional information essential for determining connectivity~\cite{genflowchart}. The conclusion is clear: connectivity extraction remains an important bottleneck in flowchart digitization, one that semantic understanding alone cannot resolve.

Recognizing these limitations, prior work on engineering diagram digitization suggests an alternative: hybrid architectures that separate element detection from connectivity reconstruction, reserving neural methods for discrete objects, while using classical techniques for line tracing~\cite{deeplearningreview,pidlines}. We implement this principle in \textit{FlowExtract}, combining YOLOv8~\cite{yolov8} for node detection and EasyOCR~\cite{easyocr} for text extraction with a novel edge detection method. Where prior approaches detect line networks first and attach directional information afterward~\cite{pidlines}, our method anchors on arrowheads as hooks for connection detection. It analyzes their orientations to determine the direction of the edges, and then traces the connecting lines backward to identify the source nodes. This design prioritizes precision over recall, proposing edges only where directional indicators have been detected rather than inferring connectivity from generic line extraction alone.

We evaluate FlowExtract on industrial troubleshooting diagrams from a consumer electronics manufacturing facility. These documents present challenges including dense technical terminology, tightly spaced nodes, and overlapping edges. FlowExtract achieves near-perfect node detection and substantially outperforms VLM baselines on edge extraction, with a precision-oriented profile where detected connections are typically correct. This profile fits well within human-in-the-loop digitization workflows~\cite{emmanouilidis2021}, where the system provides a reliable structural skeleton as a directed graph, allowing human validators to extend the extraction rather than detect and correct errors that could propagate into the final knowledge base~\cite{parasuraman2010}.

Our contributions are twofold. First, we introduce a novel arrowhead-anchored edge detection method for flowchart digitization that prioritizes precision over recall, producing directed graph representations where detected connections are reliable. Second, we provide empirical evidence that hybrid architectures separating element detection from connectivity reconstruction substantially outperform vision-language models on procedural flowcharts. Taken together, these contributions offer a starting point for organizations seeking to unlock procedural knowledge from legacy maintenance documentation, enabling its integration into modern decision support systems.

\section{Background}

\subsection{Vision-Language Model-Based Extraction}

Vision-language models have achieved strong performance on combined document-image interpretation, inferring semantically meaningful descriptions of visual content~\cite{layoutlmv3}. However, this semantic focus encounters challenges when connection topology becomes complex. Systematic evaluation on a dataset of 13,858 flowcharts, comprising both scientific diagrams and synthetic examples, reveals a consistent pattern across both closed models (GPT-4V, Claude-3, Gemini-Pro) and open models (LLaVA, Qwen-VL, InternLM, DeepSeek). While VLMs handle node and text detection reasonably well, they consistently fail to reconstruct connection topology, with edge-level F1 scores remaining below 0.30 across all tested architectures~\cite{flowlearn}. These limitations persist amid strong performance on related tasks, also requiring visual reasoning and text recognition.

Hybrid approaches, which combine the analytical abilities of VLMs or text-only LLMs with other methods that provide enhanced visual cues, also face similar limitations. A representative example uses the Segment Anything Model for zero-shot segmentation, OCR for text extraction, and GPT-3.5-Turbo for graph reconstruction from element descriptions~\cite{genflowchart}. This conversion from spatial layout to unordered textual descriptions discards the essential positional information for determining connectivity. Arrow-guided prompting takes an alternative approach by explicitly encoding arrow directions into VLM prompts~\cite{arrowvlm}, demonstrating improvements over unaugmented baselines. However, even with explicit directional cues, VLMs struggle with overlapping or densely packed arrow patterns. The common limitation across these approaches is the reliance on language models to infer connection topology from representations that encode spatial information implicitly or incompletely. Textual descriptions inherently lose the precise positional relationships of 2D layouts, and proximity becomes ambiguous when elements are described sequentially rather than spatially. 

One noteworthy exception applies the Relationformer neural architecture to jointly extract symbols and connections from piping and instrumentation diagrams, outperforming modular baselines by over 25\% in edge detection~\cite{pid2graph}. However, the studied pre-trained on over 170,000 samples and fine-tuned on tens of thousands more, with ablation studies showing catastrophic performance degradation when training data was reduced. Such data requirements are infeasible in most industrial digitization projects, where labeled diagrams are much more limited. These constraints motivate approaches that leverage the domain knowledge characteristics for methodological design, rather than learning connectivity patterns from data alone.

\subsection{Spatial Structure Extraction}

Where language models infer connectivity using their internal understanding and/or fine-tuning, alternative approaches usually exploit the geometric structure of diagrams directly. Diagrams across technical domains share a common visual grammar wherein discrete nodes are joined by connections, generally lines or arrows. This convention appears in flowcharts standardized under ISO 5807~\cite{iso5807}, in piping and instrumentation schematics, and in hand-drawn sketches alike. \textit{Nodes} can be understood as bounded regions with characteristic shapes and fixed aspect ratios. This is where neural network architectures excel, given their capacity for learned pattern matching across symbol detection and text recognition tasks~\cite{deeplearningreview}. \textit{Connecting lines}, by contrast, vary in length and orientation but can be compactly described by geometric parameters, making them amenable to line tracing techniques such as the Hough transform~\cite{hough} that follow paths through pixel space rather than recognizing learned patterns. Recognizing this asymmetry, recent systems have adopted hybrid architectures, reserving neural networks for discrete elements while relying on classical image processing techniques for line extraction~\cite{pidlines}.

Within this decomposition, different methods emphasize different elements, though arrowhead detection has emerged as a particularly effective strategy. Arrowheads can be treated as nodes themselves, since they are bounded regions with characteristic shapes amenable to pattern recognition. Work on handwritten diagram recognition has demonstrated that extending object detectors with keypoint predictors to localize arrowheads and tails as spatial coordinates can double overall edge prediction accuracy to 78.6\%~\cite{arrowrcnn}. Even the pipeline approach described in the previous section employed arrows as explicit cues for vision-language models, though without achieving comparable performance~\cite{arrowvlm}. The consistency of these results likely stems from the ubiquity of arrowheads as standardized directional cues, which justifies their use as explicit anchors from which the remainder of each connection can be inferred, thereby improving overall directed graph extraction.

\subsection{Human-in-the-loop Extraction}

\begin{figure}
    \centering
    \includegraphics[width=1\linewidth]{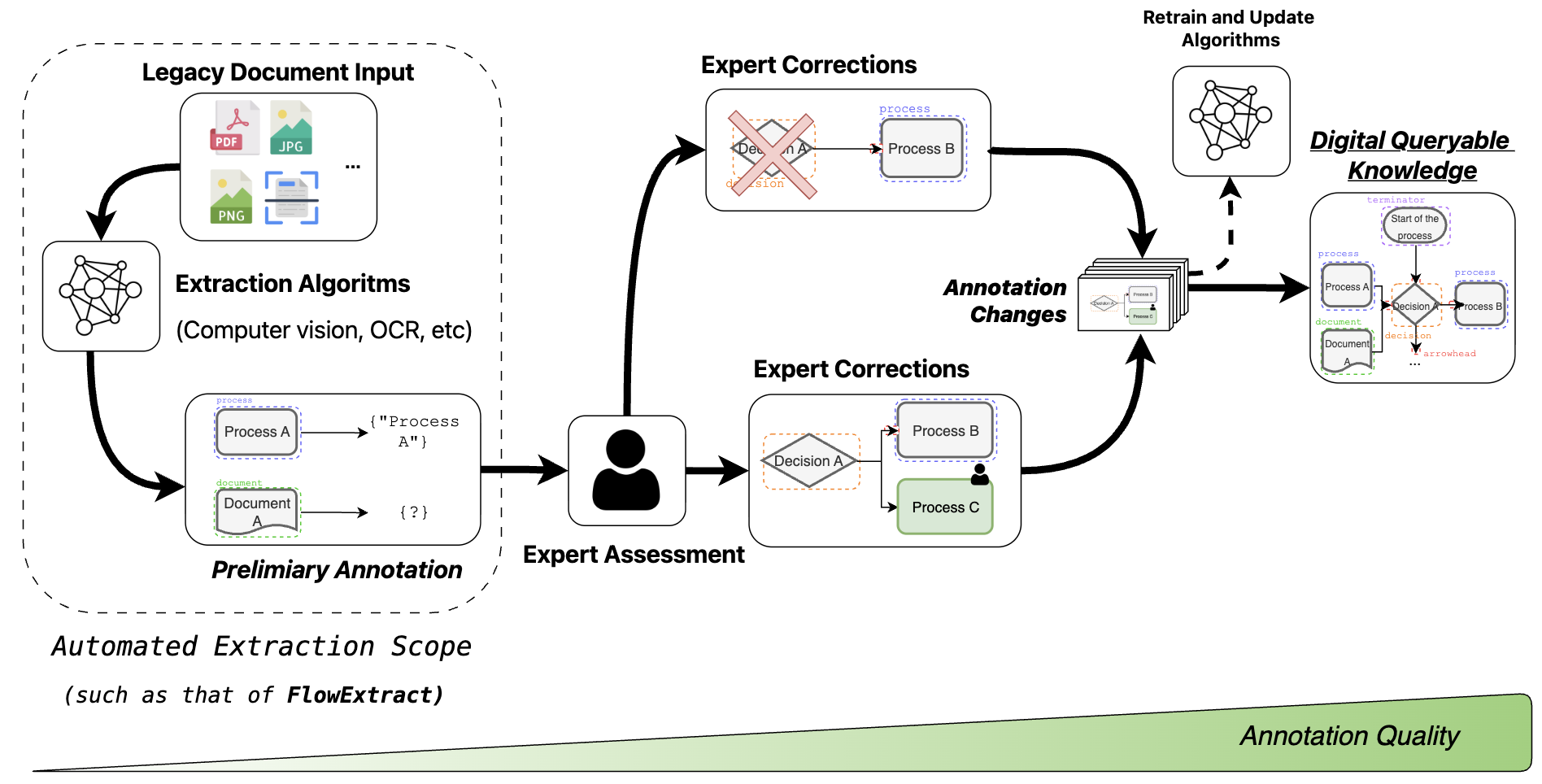}
    \caption{Human-in-the-loop workflow for knowledge extraction from legacy maintenance documentation.}
    \label{fig:hilt}
\end{figure}

Even where decomposition approaches prove more effective, they do not guarantee operational reliability. In industrial maintenance contexts, where extracted procedural knowledge informs decisions with safety consequences on top of effectiveness and efficiency requirements, fully automated extraction alone remains insufficient to obtain the data quality that would effectively mitigate such risks~\cite{emmanouilidis2019}. \textit{Human-in-the-loop} workflows address this gap by positioning domain experts that validate outputs and contribute contextual knowledge, which automated methods cannot reliably capture~\cite{proceduralkg,emmanouilidis2021}, thus maximizing the quality of extraction. Figure~\ref{fig:hilt} illustrates such workflow, in which automated extraction provides a preliminary extraction that experts correct and complete to obtain the final \textit{digital queryable knowledge} that may be used in operator support systems.

However, effective oversight of automated AI systems presents cognitive challenges that constrain how such human-in-the-loop workflows should be designed, even in the conception of the initial automated extraction. Research on \textit{automation bias}, the tendency to over-rely on automated recommendations, demonstrates that validators often accept incorrect outputs without sufficient verification~\cite{parasuraman2010}. This bias manifests itself asymmetrically. Commission errors, where validators accept incorrect outputs, prove harder to detect than omission errors, where validators notice missing information~\cite{lyell2017}. Training interventions can reduce commission errors but show limited effectiveness against omission errors~\cite{parasuraman2010}. Systems producing false positives therefore impose heavier cognitive burdens than systems producing false negatives; the former requires detecting errors embedded within plausible outputs, while the latter requires extending an incomplete but reliable skeleton. We therefore posit that extraction systems designed to fit within human-in-the-loop workflows should adopt conservative extraction strategies, prioritizing precision over recall to position validators where human cognitive abilities excel.

\section{Methodology}

\subsection{Dataset}

The present study involves a collection of troubleshooting maintenance diagrams collected from an operational consumer electronic domestic devices manufacturing facility. Using real-world operational data ensures our evaluation reflects the actual challenges of production environments, including authentic visual characteristics and domain-specific terminology. The dataset comprises 35 diagrams written in Dutch, consisting of 1145 nodes (flowchart elements, such as process boxes, decision diamonds, and terminators), edges (directional arrows connecting nodes), and decision labels (text annotations ``ja''/``nee'' in Dutch, appearing along edge paths to indicate branching logic).

\begin{table}[!t]
\caption{Class Distribution Across Dataset Splits}\label{tab:class_distribution}
\centering
\begin{tabular}{lcccc}
\hline
Class & Training & Validation & Test & Total \\
\hline
Arrowhead & 357 & 48 & 93 & 498 \\
Process & 177 & 23 & 45 & 245 \\
Document & 119 & 11 & 38 & 168 \\
Decision & 118 & 15 & 29 & 162 \\
Terminator & 35 & 4 & 9 & 48 \\
Connector & 17 & 2 & 5 & 24 \\
\hline
Total & 823 & 103 & 219 & 1145 \\
\hline
\end{tabular}
\end{table}

This dataset was manually annotated using Label Studio~\cite{labelstudio}, an open-source data labeling tool, for bounding box detection of all element types and ground truth specifying node connection and edge labels. In total, 1145 node instances were captured across all diagrams, plus an extra 62 ``ja''/``nee'' annotations. We split the dataset into 25 training images (823 instances, Table~\ref{tab:class_distribution}), 3 validation images (103 instances), and 7 test images (219 instances). The test set is held out during training and used only for final evaluation. Note that although the operational dataset is small, each image provides multiple instances of symbols, offering a somewhat richer basis for training, validation, and testing.

\subsection{FlowExtract Pipeline}

\begin{figure}[!t]
    \centering
    \includegraphics[width=\textwidth]{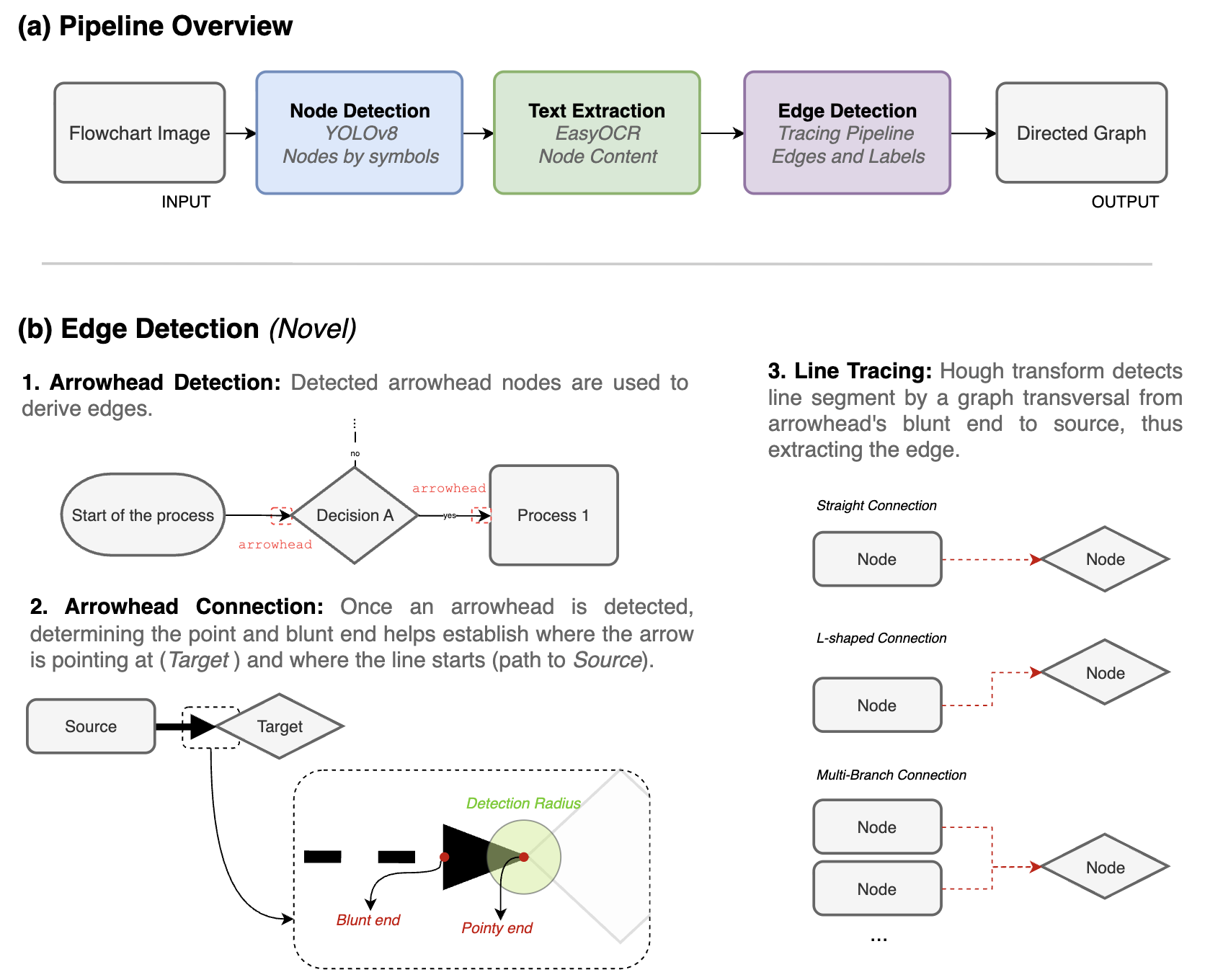}
    \caption{FlowExtract pipeline overview (a) and edge detection methodology (b), showing arrowhead orientation analysis and line tracing for straight, L-shaped, and multi-branch connections.}
    \label{fig:pipeline}
\end{figure}

Our pipeline processes troubleshooting diagram images and outputs directed graphs representing procedural flow. Fig.~\ref{fig:pipeline} illustrates the processing workflow: (a) the three-stage pipeline overview comprising node detection, text extraction, and edge detection; and (b) the edge detection methodology showing arrowhead orientation analysis and line tracing.

\subsubsection{Node Detection}
\begin{figure}[!t]
    \centering
    \includegraphics[width=0.65\linewidth]{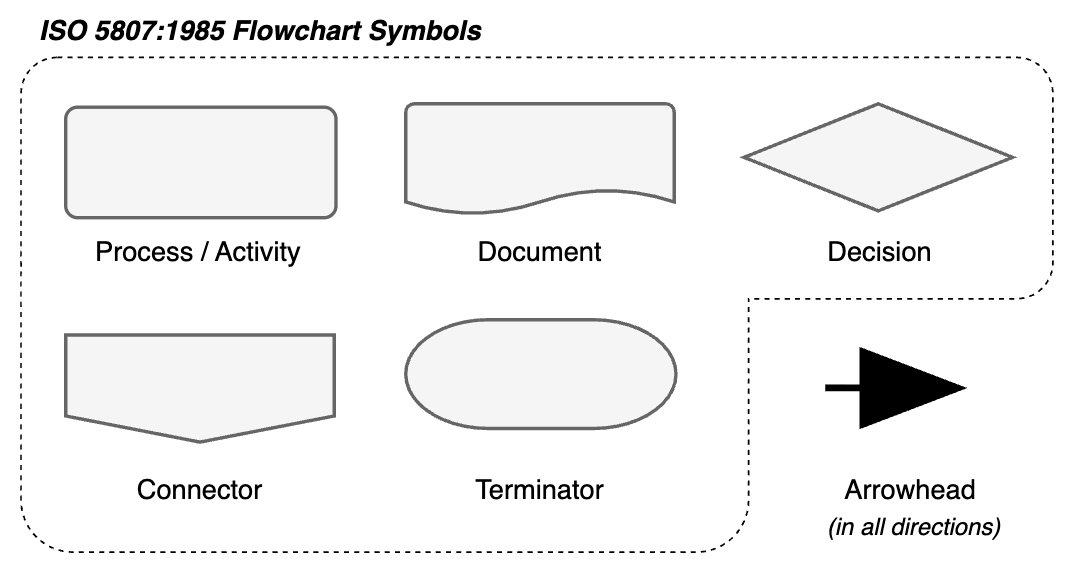}
    \caption{Flowchart symbols taxonomy (based on ISO 5807:1985) and arrowheads.}
    \label{fig:taxonomy}
\end{figure}

The first stage detects and classifies flowchart elements using YOLOv8~\cite{yolov8}, chosen for its maturity, extensive documentation, and established track record in domain adaptation that facilitates the inclusion of custom symbol sets. We define six element classes based on the subset of ISO 5807~\cite{iso5807} symbols present in our diagrams, plus arrowheads which enable subsequent edge detection, as shown in Fig.~\ref{fig:taxonomy}: Process (rectangular boxes containing action descriptions), Decision (diamond shapes containing yes/no questions), Document (rectangles with wavy bottom edges referencing external procedures), Terminator (rounded rectangles marking start/end points), Connector (circular elements with identifiers for cross-page references), and Arrowhead (arrow tips indicating edge direction, essential for edge detection but not included as graph nodes).

We initialize YOLOv8s (small variant) with weights pre-trained on the COCO dataset~\cite{coco2014}, a large-scale object detection benchmark containing 80 object categories across 330K images, providing general-purpose visual features, offering some transferability to our domain. Given the limited size of the training instances, we use the small variant to reduce overfit risk, as smaller models with fewer parameters are less prone to memorizing training examples when data is limited. However, our data set exhibits severe class imbalance, with 21 arrowheads (most common class; 357 samples) for every connector (rarest class, 24 samples), as seen in Table~\ref{tab:class_distribution}. We address this through a \textbf{Data Augmentation Strategy}, combining \textit{Mosaic augmentation}~\cite{yolov4}, which combines four training images per sample, thereby increasing our effective dataset size, with \textit{Brightness variation} (HSV, brightness metric, at $\pm$30\%) to simulate varying scan conditions. We intentionally avoid other types of augmentation (e.g., geometric, which rotates, translates, and flips the image) due to tight bounding box annotations, so as to prevent loss of spatial data characteristics.

\subsubsection{Text Extraction}
The second stage extracts textual content using EasyOCR~\cite{easyocr}, a deep learning-based OCR system that provides multilingual support and handles the text characteristics common in technical diagrams (different font sizes, orientations, and image quality of scanned documents). For each detected element (excluding arrowheads), we crop the corresponding image region and apply OCR to extract the contained text. Additionally, we perform a full-image scan to detect decision labels (``ja''/``nee'' in Dutch, corresponding to yes/no) that appear along arrow paths rather than within bounded elements. These labels are stored separately for later assignment to the edges during edge detection.

\subsubsection{Edge Detection}
The third stage reconstructs the directed graph structure by determining which elements are connected by arrows. As pointed during the Background section, we derive connections from detected arrowheads and line segments. Detected arrowhead nodes from the first stage are used to derive edges, as shown in Fig.~\ref{fig:pipeline}(b). For each detected arrowhead, we determine its orientation by identifying which end is closest to a node element: the ``pointy'' end (closest to a box) indicates the target node, while the ``blunt'' end indicates the direction toward the source. From the blunt end of each arrowhead, we trace connected line segments back toward potential source nodes by applying the Probabilistic Hough Transform~\cite{hough} to detect line segments in a preprocessed binary image. Starting from the arrowhead position, we perform a graph-based traversal of detected segments, following connected paths until reaching proximity to a node element. This approach handles straight connections, L-shaped paths, and multi-branch configurations as illustrated in Fig.~\ref{fig:pipeline}(b).

Once the source and target nodes are identified for an arrowhead, we create a directed edge. For decision nodes with multiple outgoing edges, we assign labels (ja/nee) based on the spatial relationship between detected label positions and arrow directions. Specifically, we score each label by its distance to the arrow midpoint and then run the text extractor to identify whether the label corresponds to ja or nee. Decision nodes may have multiple outgoing edges sharing a common source segment before branching; our traversal algorithm handles this by exploring all connected paths from each arrowhead independently, allowing reconstruction of complex branching structures. This arrowhead-anchored approach prioritizes precision over recall by proposing edges only where directional indicators have been explicitly detected, rather than inferring connections from line segments that may represent diagram borders, visual artifacts, or other non-connective structures.

\subsection{Training Configuration}
We train YOLOv8s for 250 epochs with image size 640$\times$640, batch size 8, and cosine annealing learning rate scheduling with initial learning rate of 0.01, following YOLOv8 default recommendations~\cite{yolov8}. The image size balances detection accuracy with computational efficiency given our hardware constraints (Apple M3 chip with 8 CPU cores and 10 GPU cores), completing training in approximately 40 minutes.

The pipeline outputs a JSON structure containing nodes (list of elements with unique ID, type, bounding box coordinates, and extracted text) and edges (list of directed connections with source ID, target ID, edge type, and optional label). This representation enables direct ingestion into knowledge constructs: nodes map to ontological concepts (e.g., process steps as procedural actions), edges represent causal or temporal relationships, and the JSON-LD format facilitates RDF conversion for semantic web integration~\cite{proceduralkg}.

\subsection{Evaluation}
We evaluate each pipeline component separately, structuring them as individual detection and classification tasks. For node detection, we measure precision, recall, and F1 for detecting the five node classes (excluding arrowheads), along with classification accuracy for assigning correct types to detected nodes. For edge detection, we measure precision, recall, and F1 for recovering directed edges between nodes, along with accuracy of assigning correct labels (ja/nee/none) to detected edges. OCR accuracy is measured as character-level match rate between extracted and ground truth text. For detection metrics, we match predicted and ground truth bounding boxes using intersection-over-union (IoU), the ratio of overlap area to union area. A detection is considered correct if IoU $> 0.5$, similar to comparable papers~\cite{pid2graph}. Higher thresholds would demand stricter localization, while lower values would risk accepting misaligned boxes.

To contextualize our results, we compare them against VLM baselines evaluated on the same set of documents in previous work~\cite{gilavalle2026procedural}. Two state-of-the-art open models were tested: Pixtral-12B and Qwen2-VL-7B. Models were prompted to extract entities (nodes) and relations (edges) from flowchart images in an instruction-guided setting, as seen in the original work~\cite{gilavalle2026procedural}. The prompt template requested structured JSON output containing node types, bounding boxes, text content, and directional connections. No token limit was imposed, and models received the full flowchart image with the extraction task description. Evaluation used the same ground truth annotations as our pipeline, with matching based on textual closeness of extracted results and category selection correctness.

\section{Results}

Table~\ref{tab:main_results} presents the main evaluation results on the held-out test set. Our proposed pipeline achieves near-perfect node detection with 98.8\% F1, compared to 34.0\% for the best VLM baseline (Table~\ref{tab:comparative_results}). The different node classes achieve detection rates above 97\%, with the exception of arrowheads, which only achieve 73.1\% accuracy. This demonstrates that YOLOv8 with mosaic augmentation effectively handles the severe class imbalance, successfully detecting rare classes (terminator, connector), amid limited training examples. Node type classification accuracy reaches 97.6\%, indicating that detection errors are primarily localization failures rather than misclassification.

\begin{table}[!t]
\centering
\caption{Detection Performance on Test Set}\label{tab:main_results}
\begin{tabular}{lcccc}
\hline
Task & Precision & Recall & F1 & Accuracy \\
\hline
Node Detection & 98.4\% & 99.2\% & 98.8\% & -- \\
Node Classification & -- & -- & -- & 97.6\% \\
Text Extraction & -- & -- & -- & 99.2\% \\
Edge Detection & 85.5\% & 54.6\% & 66.7\% & -- \\
Arrowhead Detection & -- & -- & -- & 73.1\% \\
Edge Label Assignment & -- & -- & -- & 73.8\% \\
\hline
\end{tabular}
\end{table}

\begin{table}[!t]
\caption{Results Comparison with VLM Baselines from~\cite{gilavalle2026procedural}}\label{tab:comparative_results}
\centering
\begin{tabular}{lcc}
\hline
Method & Node F1 & Edge F1 \\
\hline
Qwen2-VL-7B & 0.340 & 0.107 \\
Pixtral-12B & 0.295 & 0.015 \\
FlowExtract & \textbf{0.988} & \textbf{0.667} \\
\hline
\end{tabular}
\end{table}

For edge detection, our approach achieves 66.7\% F1 versus 10.7\% for the best VLM, representing a six-fold improvement. The results show high asymmetry between precision (85.5\%) and recall (54.6\%), indicating that when edges are detected, they are usually correct, but many ground truth edges are missed. This asymmetry reflects the arrowhead-anchored design, which proposes edges only where directional indicators have been explicitly detected rather than inferring connectivity from line structures alone. For detected edges, label assignment (ja/nee) achieves 73.8\% accuracy. Analysis reveals that edge recall is directly limited by arrowhead detection performance: since edge detection requires detecting the arrowhead that indicates connection direction, missed arrowheads directly translate to missed edges. Individual analysis reveals that missed arrowheads tend to be either small at standard zoom levels or partially occluded by overlapping line segments. Text extraction achieves 99.2\% character-level accuracy, indicating that OCR is not a significant error source; the high performance reflects the clean, printed nature of the source documents.

\begin{figure}
    \centering
    \includegraphics[width=0.8\linewidth]{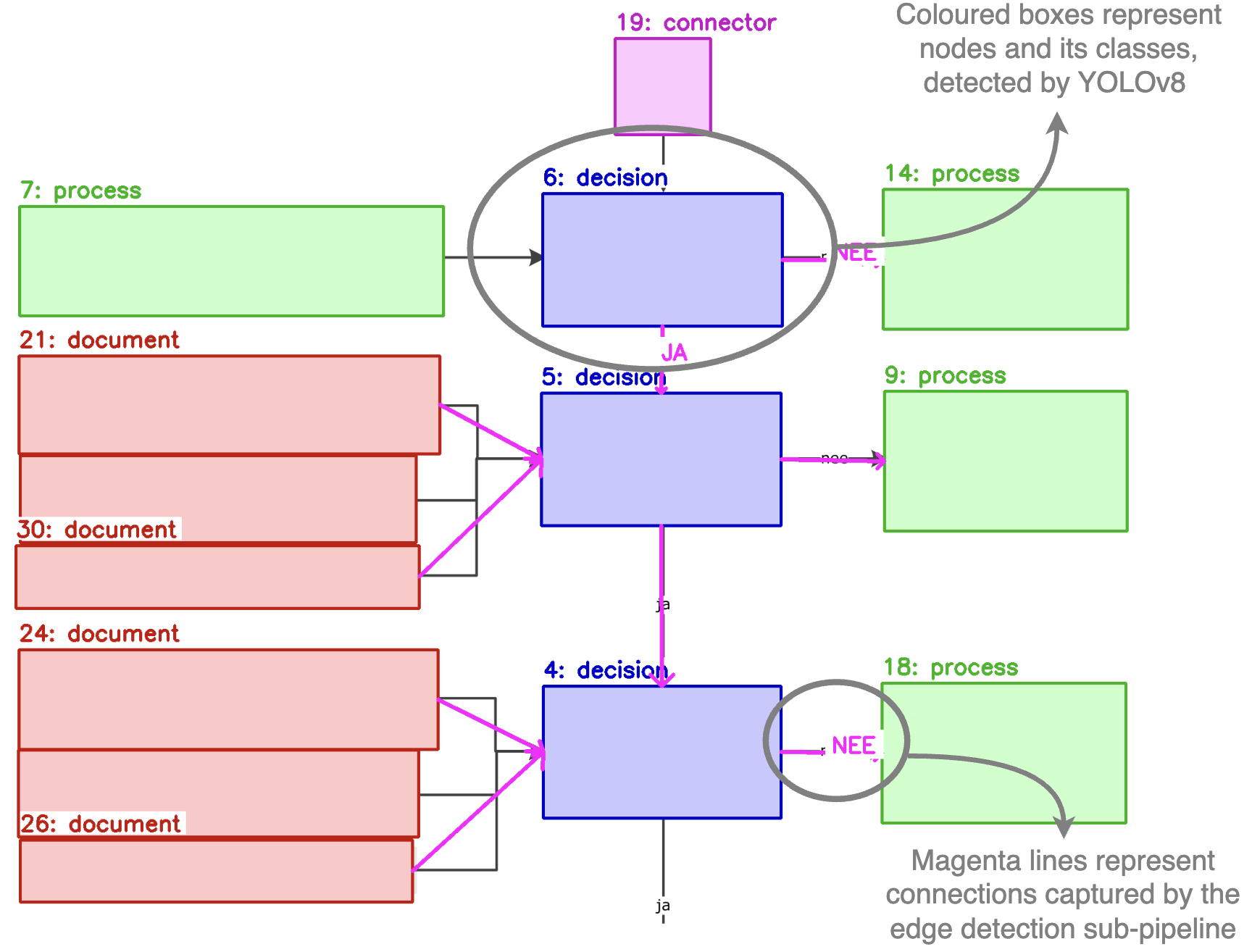}
    \caption{Example of extraction results from one of the maintenance diagrams.  The original textual content within the nodes has been computationally redacted (opaque fills) to anonymize proprietary procedural data, while preserving the structural morphology.}
    \label{fig:example}
\end{figure}

Figure~\ref{fig:example} provides a qualitative visualization of the extraction pipeline applied to a sample maintenance diagram. The visual overlay confirms the node detection stage successfully isolates discrete elements within dense layouts. More critically, the extracted connections (magenta lines) demonstrate the system's high precision to trace procedural flow, even in long documents. It also seems to handle multi-branching logic, though capturing seems to be limited to a maximum of 2-3 connections into a single node when dealing with highly dense clusters of edges, thus affecting recall. Furthermore, the figure confirms that when connections are established, the spatial heuristics accurately anchor the path to the correct source and target elements without generating hallucinatory cross-links, as seen by the fact that no magenta lines seem to fall out of the connection paths. This confirms that, while arrowhead omissions restrict total recall (such as missing branches within dense graphical clusters), the successfully parsed connections draft a precise flowchart skeleton.

\section{Discussion}

The results provide empirical support for the architectural principle identified in Section 2.2, namely that standardized technical diagrams benefit from separating element detection from connectivity reconstruction. Node detection achieves very high performance at 98.8\% F1, demonstrating that element recognition is effectively solved for this diagram class when approached with conventional object detection methods. The challenge lies entirely in reconstructing connectivity. This confirms the pattern documented across engineering diagram domains, where symbol detection and text recognition have advanced substantially through neural architectures while connection extraction remains the primary bottleneck~\cite{deeplearningreview}. Our contribution is therefore not only an ISO 5807-based diagram parser, but also targeted evidence that the separation principle transfers successfully to procedural flowcharts, with implications for how extraction systems should be architected when standardized visual conventions are present.

The 66.7\% edge F1 could reflect either a failure of structural reasoning or a failure of perception. The precision and recall split reveals the answer. Edge detection achieves 85.5\% precision but only 54.6\% recall, indicating that when arrowheads are detected, the Hough-based line tracing reliably recovers the correct source node. The bottleneck is upstream: arrowheads are missed due to small apparent size or partial occlusion by overlapping line segments. This finding is sharper than the general observation that vision-language models struggle with spatial reasoning. We have isolated the performance ceiling to a specific perceptual task rather than to the graph reconstruction logic itself. The line tracing component functions correctly; improvements to arrowhead detection translate directly into system-level gains without requiring changes to the connectivity algorithm.

The arrowheads-first architecture reflects a deliberate design choice rather than the only viable option. An alternative strategy, exemplified in recent P\&ID digitization work~\cite{pidlines}, detects the line network first and attaches directional information afterward. That approach would likely achieve higher recall, since a missed arrow does not eliminate the underlying line structure from consideration. However, lines-first introduces ambiguity in dense diagram regions where multiple lines converge toward shared junction points, requiring additional heuristics to determine which line segments constitute which edges. Arrowheads-first guarantees that each detected arrowhead corresponds to exactly one directed edge, eliminating edge-count ambiguity at the cost of recall when arrowheads go undetected. For procedural flowcharts where directionality carries semantic weight and edge identity must be unambiguous, we argue this tradeoff is appropriate. The choice should be revisited for diagram types where the underlying network matters independently of connection direction, which may not present arrowheads.

The precision-oriented performance profile directly implements the conservative extraction philosophy established in Section 2.3. High precision ensures that detected edges are trustworthy, allowing human validators to extend the extracted graph rather than correct it. This aligns with the principle of positioning human and algorithmic components where each exhibits comparative cognitive advantage~\cite{emmanouilidis2019,emmanouilidis2021}. The system contributes a reliable structural skeleton; human reviewers contribute completeness and contextual judgment that automated methods cannot reliably provide. Cognitive load during validation is reduced because reviewers are not second-guessing system outputs but rather filling gaps that the system has conservatively declined to fill. For asset lifecycle management applications where propagated errors carry operational and potential safety consequences, this division of labor supports the kind of human-AI collaboration that industrial knowledge management increasingly requires~\cite{proceduralkg,emmanouilidis2021}.

Notwithstanding the encouraging results, we also acknowledge some limitations. Firstly, arrowhead detection recall directly caps edge performance and cannot be circumvented through downstream improvements. Alternative architectures such as lines-first detection offer different tradeoffs but not unambiguous improvements; the appropriate choice depends on whether precision or recall matters more for the downstream application. A secondary limitation is scope. FlowExtract targets ISO 5807-standardized flowcharts with clean print and line work, and would require adaptation for documents employing non-standard symbol conventions (especially if not using arrows as connectors), other standards, or highly degraded documents. Finally, the pipeline processes documents in isolation and does not resolve cross-document references encoded through connector symbols. Future work could extend the system to match connector labels across documents. Merging partial graphs would enable extraction of complete procedures stemming from different documents; a common pattern in industrial maintenance where troubleshooting guides often reference sub-procedures stored separately. Assessment and validation with human operators is also among the plans for further work. 

\section{Conclusion}
We introduced \textit{FlowExtract}, a pipeline for extracting directed graphs from standardized maintenance flowcharts. The system separates element detection from connectivity reconstruction, combining standard domain-adapted node detection with a novel arrowhead-anchored edge detection method that traces connecting lines backward from detected directional indicators. Evaluated on industrial troubleshooting diagrams, FlowExtract achieves near-perfect node detection and outperforms vision-language model baselines on edge extraction by a factor of six, with a precision-oriented profile where detected connections are typically correct.

These results validate the architectural principle that standardized technical diagrams benefit from treating element detection and connectivity reconstruction as distinct tasks, with neural methods suited to bounded symbols and geometric methods suited to continuous line structures. The precision-oriented extraction profile fits well within human-in-the-loop workflows, where conservative automated extraction reduces the cognitive burden of validation by positioning human reviewers to extend an incomplete but reliable skeleton rather than detecting commission errors. For organizations seeking to digitize legacy maintenance documentation, this hybrid pipeline offers a practical starting point for unlocking procedural knowledge and integrating it into asset lifecycle management systems.

\begin{credits}
\subsubsection{\ackname} This research was funded by the European Union's Horizon Europe research and innovation programme under the AIXPERT project (Grant Agreement No. 101214389), which aims to develop an agentic, multi-layered, GenAI-powered framework for creating explainable, accountable, and transparent AI systems.  The authors wish to extend special thanks to Philips Lifestyle Consumer B.V. for the valuable collaboration and support in this work.

\subsubsection{\discintname}
The authors have no competing interests to declare that are relevant to the content of this article.
\end{credits}
%
%

\end{document}